ORIGINAL RESEARCH

**TITLE:** Automated Thematic Analyses Using LLMs: Xylazine Wound Management Social Media Chatter Use Case

**Authors:**

JaMor Hairston, MSHI MS[a] (jamor.hairston@emory.edu) (ORCID: https://orcid.org/0000-0001-6069-5869)

Ritvik Ranjan[b] (ritvikranjan@gmail.com)

Sahithi Lakamana, MS[a] (sahithi.krishnaveni.lakamana@emory.edu)

Anthony Spadaro, MD MPH[c] (avs156@njms.rutgers.edu)

Selen Bozkurt, PhD, MSc[a] (selen.bozkurt@emory.edu)

Jeanmarie Perrone, MD[d] (jeanmarie.perrone@pennmedicine.upenn.edu)

Abeed Sarker, PhD[a] (abeed@dbmi.emory.edu) (orcid: https://orcid.org/0000-0001-7358-544X)

**Author affiliations:** [a] Emory University, Atlanta, GA; [b] Wheeler High School, Marietta, GA; [c] Rutgers New Jersey Medical School, Newark, NJ; [d] Perelman School of Medicine, University of Pennsylvania, Philadelphia, PA

**Grant funding:** R01DA057599 (NIDA), 3R01DA057599-01S1 (NIDA)

**Conflicts of interest:** None

**Corresponding author:**
Abeed Sarker
101 Woodruff Circle
Office 4101
Atlanta, GA 30322

**Abstract word count:** 151 words

**Manuscript word count:** 2185 words

**References:** 14

**Figures:** 3

**Tables:** 2

**Author contributions**: JaMor Hairston – conceptualization, methodology, software, validation, data curation, writing - original draft, writing - reviewing and editing; Ritvik Ranjan – validation, writing - reviewing and editing; Sahithi Lakamana – software, validation, data curation, writing - reviewing and editing; Anthony Spadaro – conceptualization, methodology, writing - reviewing and editing; Selen Bozkurt – writing - reviewing and editing, supervision; Jeanmarie Perrone – conceptualization, methodology, writing - reviewing and editing, supervision; Abeed Sarker – conceptualization, methodology, writing - original draft, writing - reviewing and editing, supervision, funding acquisition.

**Acknowledgements**: This study was approved by Emory University Institutional Review Board (IRB) # STUDY00002458. All data used in this study were publicly available and sourced from Reddit, the social media platform, user posts.

## ABSTRACT

**Background**

Large language models (LLMs) face challenges in inductive thematic analysis, a task requiring deep interpretive and domain-specific expertise. We evaluated the feasibility of using LLMs to replicate expert-driven thematic analysis of social media data.

**Methods**

Using two temporally non-intersecting Reddit datasets on xylazine (n=286 and n=686, for model optimization and validation, respectively) with twelve expert-derived themes, we evaluated five LLMs against expert coding. We modeled the task as a series of binary classifications, rather than a single, multi-label classification, employing zero-, single-, and few-shot prompting strategies and measuring performance via accuracy, precision, recall, and $F_1$-score.

**Results**

On the validation set, GPT-4o with two-shot prompting performed best (accuracy: 90.9%; $F_1$-score: 0.71). For high-prevalence themes, model-derived thematic distributions closely mirrored expert classifications (e.g., xylazine use: 13.6% vs. 17.8%; MOUD use: 16.5% vs. 17.8%).

**Conclusions**

Our findings suggest that few-shot LLM-based approaches can automate thematic analyses, offering a scalable supplement for qualitative research.

**BACKGROUND**

The emergence of generative large language models (LLMs) has led to countless natural language processing (NLP) applications in the medical domain [1, 2]. Current LLMs demonstrate strong performance in qualitative tasks, such as synthesis and summarization, when applied to medical domain-specific texts [3, 4]. These capabilities suggest that LLMs may help scale qualitative research methodologies, which are traditionally constrained by the practical limitations of manual, resource-intensive analysis. However, LLMs remain largely untested and methodologically challenging to employ for certain qualitative tasks, such as *inductive thematic analysis.* Thematic analysis is a method for identifying, analyzing, and interpreting patterns of meaning, or themes, within data. It is typically an inductive, bottom-up process, where themes emerge directly from the data through manual analysis and interpretation [5]. Thematic analyses are often performed by experts with specialized knowledge of target domains, resulting in characterization of data through subjective, domain-specific interpretations. Automation of thematic analysis remains a methodological challenge due to the need to incorporate subjective judgment, context, and domain-specific knowledge. Additional limitations of LLMs for performing purely unsupervised thematic analyses include the tendency to perpetuate or amplify societal biases present in their training data, the inability to capture nuances of unseen data concepts, and the generation of variable results on repeat runs with identical inputs [6-8]. The latter is particularly critical for modeling thematic analysis experiments with LLMs, as large datasets, such as thousands of posts from social media corpora, cannot be processed within a single context window, and themes generated from distinct subsets of data may lack consistency.

We aim to develop and validate LLM-based strategies for replicating an expert-driven thematic analysis. Our data source, Reddit, is a popular social network that facilitates candid, anonymous discussions on sensitive topics, including substance use [9]. We specifically focus on the substance xylazine, which is an emerging concern in the ongoing opioid crisis in the United States [10]. Xylazine, an α-2 adrenergic agonist used as a sedative in veterinary care for various animal species, has increasingly been found in the illicit drug supply, rapidly expanding since 2019 [11], and has been associated with necrotizing skin wounds and lesions [12]. We detail our approach to conducting and evaluating the effectiveness of LLMs for thematic analysis of xylazine-related Reddit data, particularly via comparison against an expert-curated thematic categorization.

**MATERIALS AND METHODS**

Data collection and preparation

Data was sourced from Reddit via the PRAW (Python Reddit API Wrapper), which provides access to publicly available posts from chosen subreddits. To collect xylazine-relevant data, we employed keywords related to xylazine, as well as wound-, ulcer-, and necrosis-related terms, including common variants (*e.g.*, *tranq*) and misspellings. We divided the data into two sets based on timestamps: 2014 to 2023 (5376 posts from 961 subreddits), and 2024 to March 2025 (13,928 posts from 4272 subreddits). For manual thematic categorization, we drew random samples from each set: 286 from set 1 (DS1) and 236 from set 2 (DS2), illustrated in *Figure 1*.[1] We also retained the full filtered set 2 dataset (DS2F, 686 posts) for thematic distribution analysis. The temporal separation between datasets was intentional and used to assess generalizations across different periods, specifically because we anticipated evolution in discussions mirroring that of the opioid crisis.

Thematic analysis and manual classification

Medical toxicology-trained authors initially analyzed a random subset (n=15) of posts from DS1 to identify relevant themes based on individual post content.[2] Following theme identification and consolidation by domain experts, the larger samples from DS1 (n=286) and DS2 (n=236) were manually annotated. Twelve xylazine-relevant themes were identified as representative of the social media chatter dataset, DS1.[3] Of the classified posts that included xylazine wound-relevant content, *G:Pathophysiology of Xylazine*, *C:Location*

---

[1] We sampled 300 and 250 posts from the two sets, respectively, based on our assessment of feasibility for manual analysis, and then removed posts that were duplicative, blank, or poorly formatted posts. Illustrated in figure 1.
[2] Previously conducted work referenced in [13].
[3] Themes available in appendix.

*of Wounds*, and *L:Geography and locale* were the most prominent themes, found in 29.4%, 19.6%, and 12.9% of posts, respectively [13].

Automated thematic analysis

*Dataset 1 (DS1): Prompt development and model comparison*

Preliminary testing on DS1 revealed that LLMs performed better when the task is modeled as multiple binary classification tasks (e.g., for a given theme, 1=relevant, 0=not relevant) rather than a single multi-label classification across all 13 categories. This finding guided our adoption of the multiple binary classification framework for all subsequent experiments. We used the same manually coded subset (n=15) for iterative fine-tuning. Specifically, we adjusted prompt phrasing, task instructions, and exemplar formatting in multiple rounds of testing to align model predictions more closely with expert-coded labels and maximize F1-score.[4] Initial zero-shot experiments revealed signs of overfitting, characterized by inflated classification distributions compared to manual coding. To address this, we implemented single-shot and multi-shot prompting strategies, providing the model with representative examples of each theme.

Following prompt optimization, we scaled our evaluation to the full DS1 (n=286) using five LLMs: GPT-3.5-turbo, GPT-4o, DeepSeekV3 (deepseek-v3-0324-ud-q2_k_xl), Gemma3 (gemma-3-27b-it), and Llama3 (llama3-70b). Each model was evaluated using identical prompt structures and few-shot examples, based on precision, recall, and $F_1$ scores.

*Dataset 2 (DS2): Large-scale validation*

Based on the DS1 results demonstrating the superior performance of few-shot approaches, we deployed the two best-performing model-prompt combinations to evaluate scalability and generalizability on DS2 (n=236). The subset was used as the DS2 evaluation gold standard, and the same performance metrics used in the DS1 evaluation were applied to DS2.

---

[4] Prompts available in appendix.

Finally, we applied the established best-performing model-prompt combination to the complete DS2 filtered (DS2F) dataset (n=686). This analysis was conducted to evaluate the model's output on the larger collection of unclassified posts and to assess whether the thematic distribution patterns observed in the validation subset were consistent across the entire filtered dataset.

**RESULTS**

Prompt development and model comparison with *DS1 subset*

Table 1 summarizes the performance metrics for all model-prompt combinations evaluated on DS1. Of the zero-shot approaches, DS1_0Shot_deepseekV3 achieved the highest accuracy at 89.9%. However, its mean precision (0.611; SD=0.022) and mean recall (0.599; SD=0.022) were modest, resulting in $F_1$ score of 0.605 (SD=0.018). The other four models achieved similar results, with the closest being DS1_0Shot_gpt4o (accuracy: 89.1%; $F_1$ score: 0.567, 95%-CI: [0.528-0.604]).

The most prominent themes identified in the DS1_0Shot_deepseekV3 automation combination were *G:Pathophysiology of Xylazine* (28.7%, 82 posts), *L:Geography and locale* (18.5%, 53), and *B:Other drug use habits* (16.8%, 48). These results align well with the top relevant categories of the gold standard subset classified by subject matter experts; the LLM gold standard classes share two of their three most commonly coded themes, *G* and *L*.

In both the single- and multi-shot approaches, the GPT-4o model-prompt combinations, DS1_1Shot_gpt-4o and DS1_2Shot_gpt-4o, were identified as the best-performing, by accuracy. The single-shot approach achieved accuracy: 89.9%, 95%-CI: [88.9-90.9]; 95%-CI: [0.518-0.605]; $F_1$ score: 0.588, 95%-CI: [0.549-0.625]. The multi-shot approach achieved accuracy of 90.0%, 95%-CI: [89.0-91.0]; $F_1$ score: 0.594, 95%-CI: [0.555-0.631].

The complete theme distribution is available in *Figure 2*. The top relevant themes identified by DS1_1Shot_gpt-4o were *G* (27.6%, 79 of 286 posts), *B* (21%, 60), and *L* (18.5%, 53). The top relevant themes identified by DS1_2Shot_gpt-4o were also *G* (31.1%, 89 posts), *B* (19.9%, 57), and *L* (18.2%, 52). Similar to zero-shot approaches, these combinations share two of the three most prevalent manually-identified themes (*G* and *L*) of the gold standard subset. The DS1_2Shot_deepseekV3 model-prompt combination was a close second to DS1_2Shot_gpt-4o, with the deepseekV3 combination achieving an accuracy of 90.0% (95%-CI: [89.1-91.0]; $F_1$ score: 0.613, 95%-CI: [0.576-0.648]).

Comparison of the top two performing model-prompt combinations, using *DS2 subset*

The two top-performing model-prompt combinations (DS1_2Shot_deepseekV3 and DS1_2Shot_gpt-4o) were used to perform the automatic thematic analysis of DS2. A DS2 subset of 236 posts was manually classified as the gold standard, revealing the top themes as *B:Other drugs use habits* (40.7%, 96 posts), *H:Specific xylazine withdrawals* (26.7%, 63 posts), and a tie between *A:Xylazine use habits* and *I:Posts about MOUDs* (17.8% each, 42 posts). GPT-4o outperformed DeepSeek-V3 across all metrics, achieving 90.9% accuracy (95%-CI: [89.9-91.9]) versus 89.5% (95%-CI: [88.3-90.6]), with superior precision (0.760 vs. 0.702), recall (0.663 vs. 0.642), and $F_1$-score (0.708 vs. 0.671).

The top themes identified by DS2_gpt-4o were *B* (53%, 125), *I* (16.5%, 39), *L* (14.8%, 35), and *A* (13.5%, 32); for DS2_deepseekV3, they were *B* (43.2%, 102), *I* (19%, 45), *A* (16.1%, 38), and *L* (12.3%, 29). Three of the four top themes identified by the models matched the gold standard, and both models showed stable results across runs, with narrow confidence intervals and low variability. Their ~90% accuracy and balanced precision-recall indicate strong potential for scalable automated thematic analysis of social media content. Complete metrics and theme distributions are presented in *Table 2* and *Figure 3*, respectively.

Large-scale performance using *DS2F*

The top-performing model, DS2_gpt4o, was applied to the full filtered dataset, DS2F, of 686 posts. The distribution of this model's classified output highlighted themes *B* (65.1%, 446 posts), *I* (23.4%, 160), *L* (20.4%, 140), and *A* (14.3%, 98) as the most relevant to the posts. The xylazine wound-related themes that were least relevant to the DS2F posts were *E* (0.73%, 5), *F* (1.02%, 7), and *H* (7.45%, 51).

**DISCUSSION**

Our findings demonstrate that, with appropriate task framing (e.g., decomposing into multiple binary classification tasks rather than a single multi-label classification), systematic prompt engineering, and the inclusion of representative few-shot examples, LLMs can approximate human performance in thematic analysis. This suggests that thematic analyses can be effectively automated for large-scale data, reducing the need for expert time commitment. In the final model comparison, GPT-4o outperformed DeepSeek-V3 across all evaluation metrics, in precision (0.76 vs. 0.702), recall (0.663 vs. 0.642), and $F_1$ scores (0.708 vs. 0.671), demonstrating superiority of the closed-source GPT models over open-source DeepSeek models. However, refined prompt engineering may improve performance of current open-source models, enabling them to close the relatively small performance gap. While using API-based models, such as GPT-4o, may be costly, they can still be leveraged to reduce expert-related annotation costs. Closed-source models may be particularly appealing as they can be hosted locally, enabling the processing of large-scale data without a linear increase in cost while complying with institutional data security protocols.

Our results indicate that LLM-generated themes can track shifts in discourse over time. From DS1 to DS2, xylazine-related post volume increased, with focus shifting from wound care (DS1) to concerns about drug impurities, analogs, and fentanyl-xylazine combinations (DS2). Temporal analysis also validated our method, as many posts from 2012 to 2019 were correctly flagged as unrelated to xylazine wounds, aligning with epidemiological data showing xylazine's emergence after 2019. This demonstrates the method's ability to capture evolving discussions about substance use on platforms like Reddit. Also, this capability has broader implications for drug surveillance, as it could enable rapid detection of emerging substance use patterns and identify new trends in established substances, such as non-medical ketamine use or buprenorphine precipitated withdrawal. Such automated monitoring could provide public health officials with early warning systems for emerging drug threats by distinguishing meaningful shifts in substance use discourse from background chatter.

Our experiments revealed significant differences between zero-shot and few-shot approaches. Initial zero-shot prompts failed to capture many themes effectively, with specific categories such as *"Stigma*

*Related to xylazine wounds*" remaining completely undetected across the dataset. The implementation of single-shot and multi-shot prompt configurations improved thematic detection for most models, with the final prompts producing classification distributions that closely aligned with manual coding patterns. This suggests that while LLMs are capable of identifying themes with reliable performance, inductive thematic analysis still requires human initiation and specification of themes to facilitate LLM-driven inference. In our experiments, we attempted to test the feasibility of full automation by injecting, via the prompts, contextual information surrounding each post, including both the post content and title. However, the expanded prompts did not enhance performance and instead significantly reduced it.

Limitations

Our study and the models for thematic analysis have several limitations. Fully automated thematic analysis remains a challenge for LLMs. Even in the presence of low-shot, expert-annotated data, moderate precision and recall scores (60-76%) indicate that automated classification, while promising, still produces false positives/negatives that may adversely impact downstream analyses. Our keyword-based filtering approach may have excluded relevant content that uses alternative terminology or discusses xylazine indirectly.

Future work

Future work should focus on improving the performance of low-shot methods for thematic analysis, improving LLM agreement with human experts through iterative feedback mechanisms, and designing frameworks for conducting purely unsupervised categorization that can identify emergent themes without predetermined categories. The latter can be particularly impactful, as current unsupervised approaches, such as topic modeling, are not capable of performing expert-like thematic analyses.

**Conclusion:**

We examined the feasibility of using generative LLMs for conducting large-scale thematic analyses. Our study demonstrated that LLMs can be leveraged to conduct thematic analyses, particularly when guided by few-shot examples. Zero-shot approaches resulted in inflated estimations of high-frequency themes. Future work should explore strategies for improving LLM agreement with human experts and designing frameworks for conducting purely unsupervised categorization.

**Abbreviations:** NLP – Natural Language Processing; MOUD – Medications for Opioid Use Disorders; PWUD – People Who Use Drugs; API – Application Programming Interface

# TABLES AND FIGURES

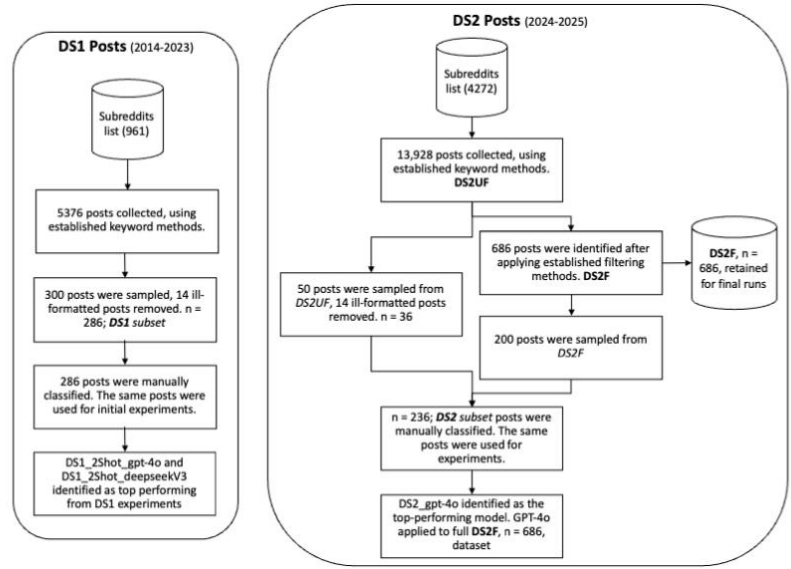

*Figure 1: Reddit Data Extraction Process*

| Model | Precision | Recall | F1 Score | Accuracy | Avg Rank |
|---|---|---|---|---|---|
| DS1_2shot_deepseekV3 | 0.614 | 0.612 | 0.613 | 0.9 | 1.625 |
| DS1_2shot_gpt-4o | 0.626 | 0.566 | 0.594 | 0.9 | 2.375 |
| DS1_0shot_deepseekV3 | 0.611 | 0.599 | 0.605 | 0.899 | 2.75 |
| DS1_1shot_gpt-4o | 0.619 | 0.559 | 0.588 | 0.899 | 3.75 |
| DS1_0shot_gpt-4o | 0.584 | 0.551 | 0.567 | 0.892 | 5.25 |
| DS1_0shot_llama3 | 0.574 | 0.478 | 0.522 | 0.887 | 7.25 |
| DS1_1shot_gemma3 | 0.522 | 0.53 | 0.526 | 0.877 | 8.0 |
| DS1_2shot_gemma3 | 0.515 | 0.568 | 0.54 | 0.875 | 8.0 |
| DS1_1shot_llama3 | 0.527 | 0.455 | 0.488 | 0.877 | 9.0 |
| DS1_1shot_gpt-35-turbo | 0.524 | 0.441 | 0.478 | 0.876 | 10.25 |
| DS1_0shot_gpt-35-turbo | 0.518 | 0.47 | 0.493 | 0.875 | 10.75 |
| DS1_2shot_gpt-35-turbo | 0.524 | 0.395 | 0.45 | 0.876 | 11.25 |
| DS1_0shot_gemma3 | 0.469 | 0.524 | 0.495 | 0.862 | 11.25 |
| DS1_2shot_llama3 | 0.511 | 0.39 | 0.443 | 0.873 | 13.5 |
| DS1_1shot_deepseekV3 | 0.354 | 0.315 | 0.333 | 0.838 | 15.0 |

*Table 1: DS1 Model Evaluations; ranked by metrics*

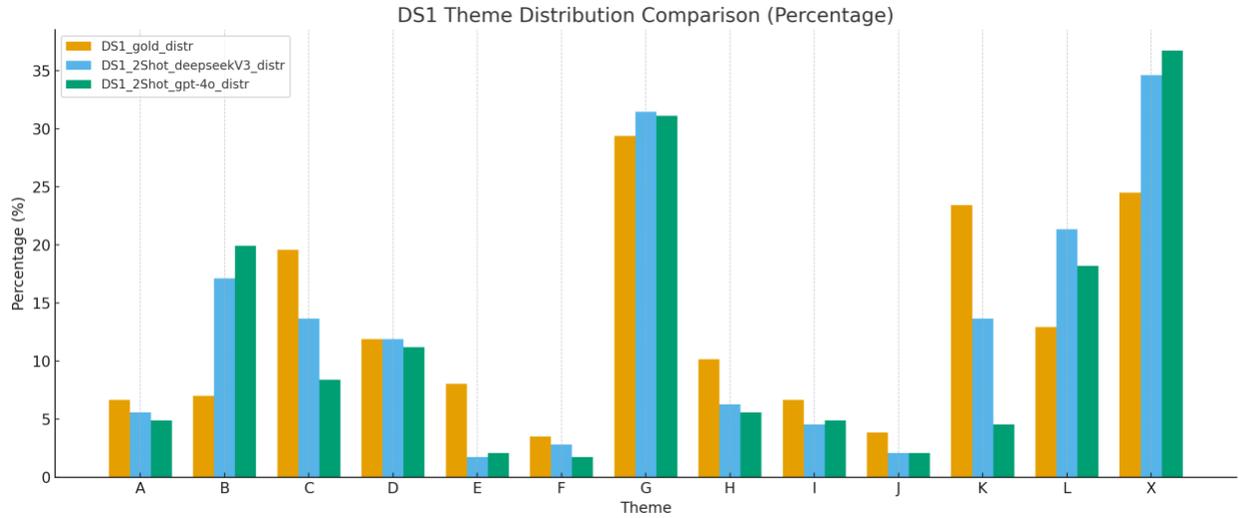

*Figure 2: DS1 top-performing models' theme distribution compared to the gold standard (out of 286 posts)*

| Model | Precision | Recall | F1 Score | Accuracy | Avg Rank |
|---|---|---|---|---|---|
| DS2_gpt-4o | 0.76 | 0.663 | 0.708 | 0.909 | 1.0 |
| DS2_deepseekV3 | 0.702 | 0.642 | 0.671 | 0.895 | 2.0 |

*Table 2: DS2 Model Evaluations; ranked by metrics*

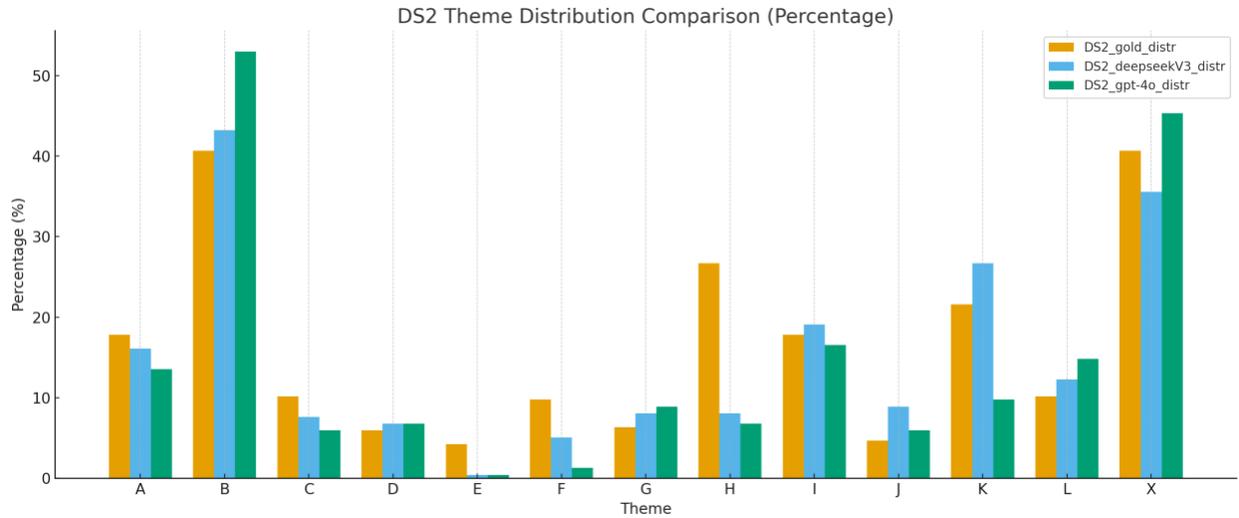

*Figure 1: DS2 models' distribution compared to the gold standard (out of 236 posts)*

# APPENDIX

1. **Prompts**

    a. Initial Prompt

    > "
    > You are a binary classification system based on raw Reddit post data. You are performing a content analysis of social media posts and determining their relevance to Xylazine-associated wounds.
    > I want you to review the following Reddit post:
    > {{post}}
    > Analyze the post's content, looking for themes and concepts as they relate to xylazine.
    > Now please answer the following questions one by one.
    > 1. Please answer if the post is relevant to 'Xylazine Use Habits'. 'Xylazine Use Habits' is defined as posts where people identify how they personally use xylazine, injection vs. snort (it should exclude people talking about how other people use it).
    > If there is relevance to 'Xylazine Use Habits', answer "1".
    > If there is no relevance to 'Xylazine Use Habits', answer "0".
    > Your answer should be formatted as "A=[answer]".
    > If a text response is generated, reanalyze the post until a 1 or 0 is generated.
    > "

    b. Major Revision

> "
> You are a binary classification system based on raw Reddit post data. You are performing a content analysis of social media posts and determining their relevance to Xylazine-associated wounds.
> I want you to review the following Reddit post:
> {{post}}
> Assess the relevance of the Reddit post to various topics related to Xylazine by answering the following prompts with "1" for relevant and "0" for not relevant.
> 1. Xylazine Use Habits: Identify if the post discusses personal use methods of Xylazine, such as injection or snorting, excluding mentions of others' use habits. Format: "A=[answer]".
> 2. Other Drugs Use Habits: Determine if the post mentions the use of other drugs alongside Xylazine, like fentanyl, oxycodone, or benzodiazepines. Format: "B=[answer]".
> .
> .
> .
> 13. Not About Xylazine: Determine if the post doesn't mention Xylazine in any context, indicating irrelevance to the topics above. Format: "X=[answer]".
> Ensure each question is answered with "1" for relevance or "0" for non-relevance. Compile all responses in a single line, separated by ', '. If a question doesn't initially yield a binary response, reanalyze until one is provided.
> "

c. Final

> "
> Task: You are a binary classification system analyzing raw Reddit posts for relevance to Xylazine (commonly called tranq)-associated wounds and use. Your job is to assess the relevance of this post to 13 specific topics. For each topic, answer with "1" if relevant or "0" if not. Format each answer as specified (e.g., A=1, B=0, etc.). Output all answers on a single line, separated by commas.
> - Review the Reddit post below.
> - For each category (A through X), output "1" if the post is relevant or "0" if it is not.
> - Output the results as a single line, strictly in the format:
>   A=_, B=_, C=_, D=_, E=_, F=_, G=_, H=_, I=_, J=_, K=_, L=_, X=_
> - Do not include explanations in the output. Only the classification line.
> Reddit Post:
> {{post}}
> Relevance Categories:
> A. Xylazine Use Habits
> Does the post describe the author's personal use of Xylazine (e.g., injecting or snorting)? Exclude mentions of others' use.
> .
> .
> .
> L. Geography or Locale
> Does it mention specific cities, states, or regions where Xylazine is found or used?
> X. Not About Xylazine
> Is the post entirely unrelated to Xylazine (i.e., no mention at all)?
> Examples:
> 1. Post:
> "Yes clonidine is the best drug I'd say for coming off tranq... I wanted to scratch off my own skin... blocks NPI release making you feel totally calm..."
> Classification:
> A=0, B=1, C=0, D=0, E=0, F=0, G=1, H=1, I=0, J=1, K=0, L=0, X=0
> .
> .
> .
> ..
> "

2. **Themes**

   *A: Xylazine Use Habits:* Posts where people identify how they personally use xylazine, injection vs snort (should exclude people talking about how other people use

*B: Other Drugs use habits:* Posts where people mention what other drugs they use, fent, oxy, benzos, etc

*C: Locations of wounds:* Posts that describe where wounds occur on the body, looking for mouth, nose, extremities, or other locations

*D: Management of Wounds:* Posts about what people are doing to manage wounds, dressings, changing, medical interventions like antibiotics, hospital admission, amputations

*E: Stigma Related to Xylazine wounds:* Posts about stigma, or posts using stigmatizing language, i.e., zombie, flesh-eating, apocalypse

*F: Ability to get into rehab clinics:* Posts about being able to get into rehab clinics, xylazine wound-related or not.

*G: Pathophysiology of Xylazine:* Posts trying to explain why xylazine does what it does, looking for comparisons to other drugs, like krokodil and clonidine, or alpha effects

*H: Posts about specific xylazine withdrawal symptoms:* Any posts that mention xylazine withdrawal specifically, would exclude ones that don't specify substance withdrawing from

*I: Posts about MOUDs:* Any posts that mention methadone, buprenorphine, subs, etc

*J: Non-MOUD management of withdrawal:* Posts about the management of xylazine withdrawal symptoms with non-MOUD medications like gabapentin, benzos, clonidine.

*K: Non-relevant post about xylazine:* A post that mentions xylazine but doesn't seem to be about personal use, withdrawal, or wounds.

*L: Geography and locale:* Posts that specifically include geographical demographic information. Looking for specific mentions of location and presence of xylazine in the area

*X: Not about xylazine at all:* A post that has nothing to do with xylazine, wounds, withdrawal